\documentclass[letterpaper, 10 pt, conference]{ieeeconf}

\IEEEoverridecommandlockouts                              % This command is only needed if 
\usepackage{amsmath} % assumes amsmath 
\usepackage{float}
\usepackage{mathtools}
\usepackage{breqn}
\usepackage{bm}
\usepackage{ushort}
\usepackage{mathtools}
\usepackage{svg}
\usepackage{pdfpages}
\usepackage{graphicx}
\usepackage{cite}
\usepackage{amssymb,amsfonts}
\usepackage{algorithmic}
\usepackage{textcomp}
\usepackage{xcolor}
\usepackage{hyperref}
\usepackage{subfig}
\newtheorem{definition}{Definition}[section]

\title{\LARGE \bf
Safe Planning for Self-Driving Via Adaptive Constrained ILQR}

\author{Yanjun Pan$^{1}$, Qin Lin$^{2}$, Het Shah$^{3}$, and John M. Dolan$^{2}$% <-this % stops a space
\thanks{$^{1}$Yanjun Pan is with Department of Mechanical Engineering, Carnegie Mellon University, 
5000 Forbes Ave, Pittsburgh, PA 15213, USA {\tt\small yanjunp@andrew.cmu.edu}}%
\thanks{$^{2}$Qin Lin and John Dolan are with Department of the Robotics Institute, Carnegie Mellon University, 
        5000 Forbes Ave, Pittsburgh, PA 15213, USA {\tt\small qinlin,jdolan@andrew.cmu.edu}}
\thanks{$^{3}$Het Shah is with Department of Mechanical Engineering, Indian Institute of Technology Kharagpur,
        Kharagpur, West Bengal 721302, India}%
}

\begin{document}

\maketitle
\thispagestyle{empty}
\pagestyle{empty}

%%%%%%%%%%%%%%%%%%%%%%%%%%%%%%%%%%%%%%%%%%%%%%%%%%%%%%%%%%%%%%%%%%%%%%%%%%%%%%%%
\begin{abstract}
%Autonomous vehicles will encounter different scenarios when driving on a high-speed freeway or on a slow-speed local avenue. Such scenarios require different motion planning approaches. 

Constrained Iterative Linear Quadratic Regulator (CILQR), a variant of ILQR, has been recently proposed for motion planning problems of autonomous vehicles to deal with constraints such as obstacle avoidance and reference tracking. However, the previous work considers either deterministic trajectories or persistent prediction for target dynamical obstacles. The other drawback is lack of generality - it requires manual weight tuning for different scenarios. In this paper, two significant improvements are achieved. Firstly, a two-stage uncertainty-aware prediction is proposed. The short-term prediction with safety guarantee based on reachability analysis is responsible for dealing with extreme maneuvers conducted by target vehicles. The long-term prediction leveraging an adaptive least square filter  preserves the long-term optimality of the planned trajectory since using reachability only for long-term prediction is too pessimistic and makes the planner over-conservative. Secondly, to allow a wider coverage over different scenarios and to avoid tedious parameter tuning case by case, this paper designs a scenario-based analytical function taking the states from the ego vehicle and the target vehicle as input, and carrying weights of a cost function as output. It allows the ego vehicle to execute multiple behaviors (such as lane-keeping and overtaking) under a single planner. We demonstrate safety, effectiveness, and real-time performance of the proposed planner in simulations.%Results from different test cases demonstrate satisfactory behavior of such a planner over a mixture of lane keeping and overtaking scenarios.

\end{abstract}

%%%%%%%%%%%%%%%%%%%%%%%%%%%%%%%%%%%%%%%%%%%%%%%%%%%%%%%%%%%%%%%%%%%%%%%%%%%%%%%%
\section{INTRODUCTION}
On-road autonomous vehicles require obstacle motion prediction to efficiently plan for future trajectories. Though motion planning has been extensively studied in recent decades, safe planning with assurance in highly uncertain environments is still challenging, especially for safety-critical systems such as autonomous vehicles. One problem arises from the highly uncertain trajectories of surrounding vehicles due to sensing, localization, maneuver or intention uncertainties, etc. Probabilistic approaches using Extend Kalman Filters (EKFs) or Unscented Kalman Filters (UKFs) can be leveraged for the state estimation and prediction for target vehicles. However, they work best on the assumption of a Gaussian distribution, which does not always hold in practice. Although particle filters are proven effective against arbitrary distributions, determining the number of particles required to guarantee a convergence is tricky and sometimes computationally heavy. More importantly, they are only capable of providing statistical guarantees in belief space. Constrained Iterative Linear Quadratic Regulator (CILQR) has been proposed for collision avoidance for autonomous vehicles \cite{cILQR,cILQR2}. Unfortunately, previous work only demonstrates the validation of the optimization framework by considering deterministic trajectories or persistent prediction for target vehicles. \emph{The first research question raised in this study is how to incorporate a sophisticated prediction for target vehicles into Iterative Linear Quadratic Regulator (ILQR) framework and assure safe planning considering prediction uncertainty.}

The second challenge the existing CILQR framework faces is the tuning of the cost function weights. Cost function weights heavily influence planned trajectory. For instance, if the reference tracking term in the cost function dominates the velocity term, a lane keeping behavior behind a slow vehicle is encouraged. If the dominance reverses, overtaking behavior is encouraged instead. Unfortunately, there is no study about the weight tuning problem in CILQR. \emph{The second research question we aim to address is how to design a scenario-aware mechanism to adaptively generate appropriate weights for diverse scenarios.}

For the first question, we use one reachability analysis to enclose all kinematically possible trajectories of a target vehicle into an envelope represented by a set considering the uncertainty of its initial states (position, yaw, etc.). We allow the target vehicle to behave non-determistically over the short-term prediction horizon. Using this envelope, a motion planner can safely avoid all possible extreme scenarios in short-term. However, as the size of the envelope grows with time, it might block the entire traversable path, which explains why it has been only used in the short-term prediction. To avoid being overly conservative and avoid a local optimum, an adaptive least square filter is used for long-term predictor. This part can be easily substituted by more complicated and more accurate predictions. Our optimization framework solves the collision avoidance problem in terms of occupancy from the short-term and the long-term prediction in a unified manner. To address the second question, we design a scenario-based analytical function taking the states from the ego vehicle and the target vehicle as input, and carrying weights of a cost function as output. It allows the ego vehicle to execute multiple behaviors (such as lane-keeping and overtaking) under a single planner. %This particular function allows the ego vehicle to perform lane keeping when a forward vehicle is travelling at an adequate velocity and it allows the ego vehicle to perform an overtake maneuver when a slow forward vehicle does not yield to the ego vehicle under the same planner.

Our work makes the following contributions:
\begin{enumerate}
    \item We incorporate a hybrid prediction crossing short-term and long-term levels for a safety-assured and efficient planner under highly uncertain surrounding traffic environments.
    \item We propose an analytical function based on ego and target vehicle state difference and target vehicle velocity. This function increases the generality of a single planner under the same set of cost function weights.
    \item The whole framework has achieved real-time performance with average cost around 65ms over a 4s prediction horizon in simulations with a 0.1-second discretization step.
\end{enumerate}
The rest of this paper is organized as follows. Section \ref{sec:related_work} provides a review of some important related work regarding motion planning methods and predictions. Section \ref{sec:method} is the detailed framework proposed in this paper and necessary related background materials. Section \ref{sec:problem_def} is about the problem formulation relating to the cost function used and vehicle modeling. Section \ref{sec:result} presents the experimental results. The conclusions and the future work are in Section \ref{sec:conclusion}.
 %A function that relates ego vehicle and target vehicle states to cost function weights is proposed and tuned. 
%%%%%%%%%%%%%%%%%%%%%%%%%%%%%%%%%%%%%%%%%%%%%%%%%%%%%%%%%%%%%%%%%%%%%%%%%%%%%%%%
\section{RELATED WORK} \label{sec:related_work}
% There are three major categories of classical planning algorithms for autonomous vehicles: search-based method, sampling-based method, and optimization-based method\cite{gonzalez2015review}. Search-based methods (such as A* and Dijkstra) are generally efficient for dealing with non-dynamic environments and are globally optimal. Sampling-based methods (such as RRT* \cite{RRT_star}) provide an efficient way to traverse a high-dimensional state-space; they often require further trajectory smoothing and their random nature prevents them from having a set convergence rate. Optimization-based methods formulate the motion planning problem as a numerical optimization problem with a smooth trajectory and control sequence as outputs. This paper falls into the optimization-based category.

% There are three major categories of classical planning algorithms for autonomous vehicles: search-based method such as A* and Dijkstra \cite{gonzalez2015review}, sampling-based method, and optimization-based method. Search-based methods, such as RRT* \cite{RRT_star}), and optimization-based methods. This paper falls into the optimization-based category.

%The problem of trajectory planning and tracking through control of vehicle steering and acceleration has been widely studied.
\subsection{Trajectory optimization-related}
Trajectory optimization can be formulated as a constrained quadratic program\cite{rw1}. Ziegler et al. \cite{benz} used Sequential Quadratic Programming (SQP) to solve the nonlinear and non-convex problem, but the computation time is around 0.5s, which is not suitable for real-time implementation. Xu et al. \cite{rw3} proposed an efficient real-time motion planner with trajectory optimization that discretizes the plan space and selects the best trajectory based on a cost function.

Control methods such as Model Predictive Control (MPC) have also been used for this purpose. Two-layer MPC \cite{2mpc}, which reduces the computation cost as compared to one-layer MPC, has been proposed for motion planning with obstacle avoidance. Arab et al. \cite{arab2016motion} presented a Sparse-RRT* and nonlinear MPC-based motion planner for aggressive maneuvers. Ji et al. \cite{mmpc} present a path planning and tracking framework using 3D potential fields and multi-constrained MPC. Borrelli et al. \cite{borrelli2005mpc} propose a nonlinear MPC to stabilize the vehicle along a desired path, but there is a trade-off between planning speed and the prediction horizon. Heavy computation still stands as a barrier to nonlinear MPC. 
\subsection{ILQR-related}
Iterative linear quadratic regulator (ILQR) is an optimization-based method for nonlinear systems with lower computation time by utilizing dynamic programming. However, very few works have considered constraints in ILQR. Control-limited differential dynamic programming (DDP), proposed in \cite{lqr1}, deals with constraints of upper and lower bound on control inputs. Extended LQR \cite{lqr2} - \cite{lqr3} works with collision avoidance for circular obstacles, penalizing the distance from the center of the obstacle. Chen et al. proposed a constrained ILQR for autonomous vehicle motion planning \cite{cILQR}. It includes an elliptical over-fitting for obstacles and utilizes barrier functions to incorporate various constraints into the cost function. However, test cases do not demonstrate its capability of handling target vehicles with lateral motion, nor can it use the same set of weights to perform both overtaking and lane keeping behaviors over a wide variety of scenarios. Chen et al. \cite{cILQR2} proposed an improved barrier function to impose hard constraints on optimization process (i.e., the ego vehicle will not traverse into unfeasible regions). However, it still does not solve the problem of weight-tuning. Another significant drawback in \cite{cILQR,cILQR2} is that they only consider deterministic trajectories or persistent prediction for target vehicles.
\subsection{Prediction-related}
Readers are referred to a comprehensive literature review about trajectory prediction in Ref. \cite{lefevre2014survey}. Althoff et al., perform an online verification of a planned trajectory using reachability analysis \cite{althoff2014online}, i.e., ``planning-then-verification". They propagate the reachability of an ego vehicle and its surrounding vehicles simultaneously over the whole planning horizon to verify the overlap. A re-planning step is needed if the path is verified to be unsafe. However, in our framework, the reachability has been plugged into the optimization, i.e., the planned trajectory has already been guaranteed to be safe to execute without re-planning. In addition, we argue for the use of a short-term horizon instead of a whole planning horizon for safety to avoid over-conservative planning. We get inspired by the spirit of combing short-term and long-term planning in Ref. \cite{liu2018robot}. They use two-layer planners, whereas, we propose to use a single planner to combine short-term and long-term planning to a unified framework to improve the efficiency. The details of analyzing the difference will be discussed in the methodology section. 

\section{METHODOLOGY}\label{sec:method}
In this section, we discussing the sequential framework containing a mixture of short- and long-term target trajectory predictions using reachability analysis and an adaptive constrained ILQR optimization-based motion planner. 

\subsection{Prediction of dynamic obstacles}
\begin{figure}[h]
\centering
\includegraphics[width=8cm]{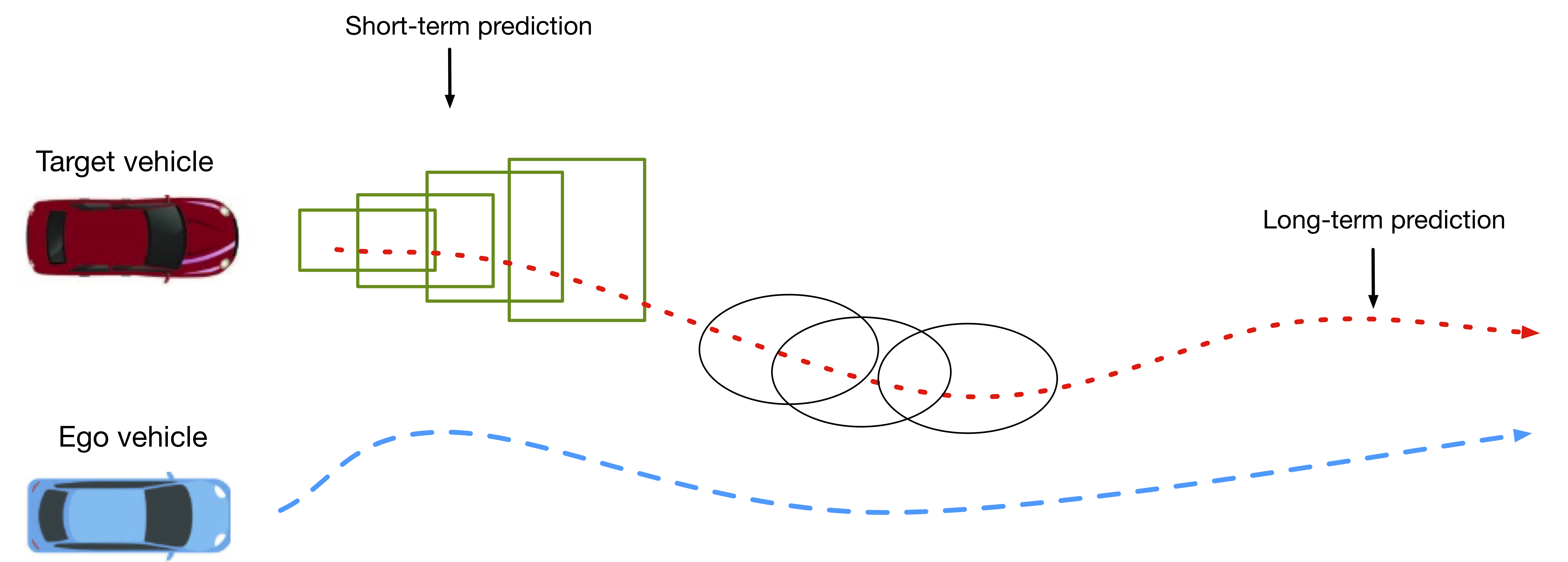}
\caption{High-level illustration for short-term and long-term prediction for planning. For a 4s predictive horizon, we use reachability in the first 0.5s, and an adaptive filter in the remaining 3.5s for prediction, respectively. The sampling time is 0.1s, thus we will have 40 ellipses as obstacles for collision avoidance.}
\label{fig:highlevel}
\end{figure}

The idea of hybrid prediction and planning is inspired by Liu et. al \cite{liu2018robot}. The authors proposed to combine a safety-oriented short-term planner and an efficiency-oriented long-term planner to deal with a safe human-robot interaction problem. The short-term planner considers the prediction uncertainty of the human trajectory, which is critical for safety. However, the disadvantage is that the uncertainty will propagate over time to make the planner excessively conservative. The prediction uncertainty is not considered in the long-term planner for the sake of efficiency. In every iteration, the overlap between long-term planned trajectory and short-term reachable region is checked. The long-term planning will be executed if no interaction is detected, i.e., the long-term planner is already safe. Otherwise they switch to the short-term planner. To further improve the efficiency, we propose to optimize the short-term and long-term collision avoidance at the same time in a unified framework without having two separate planners and extra intersection checking. A high-level illustration is shown in Figure \ref{fig:highlevel}. The short-term prediction considers the uncertainty of the target vehicle's initial state (e.g., perception error for the localization) and the uncertainty of its control actions over the short-term prediction horizon under a kinematically feasible but possibly non-deterministic assumption. The reachable positions are illustrated as rectangles. Note that they will be further inflated into ellipses considering the vehicle's shape. The long-term prediction does not consider the uncertainty. Each prediction mass point position will also be inflated into an ellipse. Another significant difference is that they use a tailored convex optimization framework by leveraging safe set and convex feasible set.

\subsubsection{Short-term prediction with reachability analysis}
In this paper we use a prototype continuous model checker based on the library of Flow* \cite{chen2013flow}, which is a nonlinear reachability analysis tool. Given the uncertainty of states and a prediction horizon, our tool is able to compute the reachable states. All possible future trajectories will fall within the reachable region with theoretical guarantee, since the reachability is an over-approximation of dynamics.

The dynamics of a plant are defined as ordinary differential equations (ODEs), i.e., $\frac{d x}{dt} = F(x, t)$. The reachability analysis essentially tries to solve an \emph{initial value problem}, i.e., given the interval of an initial state $x_0 = [\ushort{x}_0, \bar{x}_0]\in \mathcal{X}_0 \in \mathcal{I}^n$ at time 0 and a time interval $[0, t]$, we compute the enclosure $\mathcal{X}_{[0, t]}$ as the reachable states from $\mathcal{X}_0$ over $[0, t]$. However, we need to solve for the explicit primitive function $f$ of $F$, which is only possible for simple ODEs. Instead, Flow* does not solve for the explicit form of $f$. Iterative algorithms can be used to derive a Taylor model consisting of polynomial and safe remainder to over-approximate $f$; the details of implementation can be found in Ref. \cite{chen2012taylor}.

\begin{definition}{\emph{(Taylor Model)}:}
A Taylor Model (TM) of order $k$ over domain $D$ is denoted by a pair $(p,\mathcal{I})$, wherein $p$ is a Taylor polynomial of degree at most $k$, and $I$ is a \emph{remainder interval}. Given a TM $(p,\mathcal{I})$ and a function $f$ which are both over the same domain $D$, $f$ is over-approximated by $(p, \mathcal{I})$, denoted by $f \in (p, \mathcal{I})$, i.e., $f(x) \in p(x) +\mathbb{I}, \forall {x} \in D$.
\end{definition}

\begin{definition}{\emph{(Taylor Polynomial)}}
Given a $K$ times differentiable multivariate function $f \in \mathcal{C}^\kappa(D)$, where $\mathcal{C}$ is set of functions, the domain $D\subseteq \mathbb{R}$, the $k$-order Taylor approximation for $k \leq K$ that expands at the center $x = c$ for $c \in D$ is called a Taylor polynomial:

\begin{equation}
\begin{split}
    p_k(x) &= f(c) + \sum_{i=1}^n \left( \frac{\partial f}{\partial x_i}(c) \cdot (x_i-c_i)\right) + \cdots \\
    & + \frac{1}{k!} \sum_{j_1+\cdots+j_n=k} \left( \frac{\partial^k f}{\partial x_1^{j_1} \cdots \partial x_n^{j_n}}(c) \cdot \prod_{i=1}^n(x_i-c_i)^{j_i} \right)
\end{split}
\end{equation}
\end{definition}
In this work, we consider the uncertainty of initial states of the target vehicle's $x$ position, $y$ position, and yaw. Their initial bounds are user-defined parameters due to perception errors. Note that the future control actions for the target vehicle over the prediction horizon are not observable. To ensure the safety guarantee, we augment ODEs for the control actions, i.e., $\dot{v} = \mathcal{I}_v$ and $\dot{\delta} = \mathcal{I}_{\delta}$. The upper bound and the lower bound set in our paper are $\mathcal{I}_v = [-4, 6]$ and $\mathcal{I}_{\delta} = [-0.1, 0.1]$ \cite{zhang2018lane}. The interpretation for the velocity interval is that the maximum deceleration and acceleration are $-4\ m/s^2$ and $6\ m/s^2$ respectively. The $\delta$ interval is the maximal change in the steering angle within a second. Non-deterministic behaviors of the target vehicle are allowed in deriving the enclosure of the reachable region.
\subsubsection{Long-term prediction with recursive least squares filter}
In this paper, we use an adaptive linear autonomous dynamical model: $\dot{x} = A_{t} x$, where $x = [p_x, p_y, \theta]^T$. $A_t$ is a time-variant parameter matrix. With the new observation of a true state, $A_t$ is updated accordingly. The cost function and its gradient form are as follows:
\begin{equation}
\min _{\zeta} L_{k}=\frac{1}{2}\left\|x_{k}-f_{\zeta}\left(x_{k-1}\right)\right\|^{2}+\lambda L_{k-1}
\end{equation}
\vspace{-0.2cm}
\begin{equation}
    \nabla_{\zeta} L_{k}=-\nabla_{\zeta} f_{\zeta}\left(x_{k-1}\right)^{T}\left[x_{k}-f_{\zeta}\left(x_{k-1}\right)\right]+\lambda \nabla_{\zeta} L_{k-1}=0
\end{equation}
where $\zeta$ is a vector containing the parameters in $A_t$ we want to optimize, $f$ is the predictive model, and $\lambda$ is the forgetting factor, which can be tuned as an adaptive rate.
The explicit update form can be found in the literature \cite{goodwin2014adaptive,liu2015safe}. The advantage of such a linear filter is its efficiency and adaption ability. The disadvantage is that kinematic constraints, road geometry, and drivers' intentions are not considered. Sophisticated long-term prediction methods such as LSTM and RMIN \cite{Dong2019RMIN} network can be used to replace it. However, considering the real-time demand of the whole framework and the purpose of using long-term prediction to avoid the local optimum problem of short-term planning, the linear adaptive filter approach is applied in our work.
% \begin{equation}
%     \theta_{k}=\theta_{k-1}+\left[G_{k-1}^{T} G_{k-1}+\lambda H_{k-1}\right]^{-1} G_{k-1}^{T} e_{k}
% \end{equation}

\subsection{Iterative Linear Quadratic Regulator}
Autonomous vehicle obstacle-free motion planning (i.e., lane-keeping) can be formulated as a standard ILQR problem with nonlinear system dynamics:

\begin{align}
& \underset{x, u}{\text{min}}
& J = x_N^T Q_N x_N + x_N^T p_N x_N + q_N \nonumber\\
& &  + \sum_{t=0}^{N-1} (x_t^T Q_t x_t + x_k^T p + u_t^T R_t u_t) \label{eqn:ILQR_obj_fun}\\
& \text{s.t.}
& x_{t+1} = f(x_t, u_t) \label{eqn:ILQR_const}
\end{align}
where $x_t$ and $u_t$ are the state and the control input at time step $t$. Equation \ref{eqn:ILQR_const} is the system dynamics constraint, which is a transition function mapping state and control at step $t$ to state at step $t+1$. 

Since the standard LQR only solves optimization problems with quadratic cost and linear systematic constraints, this problem can be reformulated. By linearizing the systematic constraint at multiple points, we can relax the nonlinearity of the ILQR problem into the linear problem required by LQR. The steps are listed below.
\begin{enumerate}
    \item Start with a feasible initial guess $\hat{u_t} = u_0$ and obtain $\hat{x_t} = x_0$ using the system dynamics constraint. If no feasible initial guesses are available, all zero initializations with small perturbations are feasible as well.
    
    \item Calculate the derivatives of the dynamics and the cost function about the trajectory.
    \begin{gather}
    F_t = \nabla_{x_{t}, u_{t}}f({\hat{x_{t}}, \hat{u_{t}}}) \\  
    l_t = \nabla_{x_{t}, u_{t}}J({\hat{x_{t}}, \hat{u_{t}}}) \\
    L_t = \nabla^2_{x_{t}, u_{t}}J({\hat{x_{t}}, \hat{u_{t}}})  
    \end{gather}
    
    \item Run LQR backward pass on state $\delta x_t = x_t - \hat{x_t}$ and action $\delta u_t = u_t - \hat{u_t}$.
    \item Run forward pass with real nonlinear dynamics.
    \item Update $\hat{x_t}$ and $\hat{u_t}$ based on states and actions in forward pass.
    \item Iterate the whole process until the cost value converges. 
\end{enumerate}

%The detailed procedure can be found in \cite{ILQR}.

Although dynamic programming provides an efficient solution for ILQR, ILQR itself has the drawback of its constraint-free nature and it also requires the cost function to be quadratic. The constraint-free nature makes it insufficient for direct application to the autonomous vehicle motion planning problem. In the meantime, the quadratic requirement can be full-filled by applying Taylor Approximation.

\subsection{Constrained Iterative Linear Quadratic Regulator}
The constrained ILQR (CILQR) algorithm offers the inclusion of different constraints into the objective function through barrier functions. Constraints can be generalized into two categories by linearity. First, any nonlinear constraints can be converted to linear constraints via a second-order Taylor Expansion. Then, a barrier function is applied and quadratized. Equation \ref{eqn:bf} and Equation \ref{eqn:bf_quad} demonstrate this process. The quadratized linear barrier function can now be incorporated into the ILQR algorithm. 

An exponential barrier function is chosen, as it has easy-to-derive analytic derivatives in contrast to use of a high-time-complexity finite difference method. The detailed algorithm can be found in \cite{cILQR}.
\begin{equation}
    b_k^x = q_1 exp(q_2 f_k^x)
    \label{eqn:bf}
\end{equation}
\begin{multline}
    b_k^x(x_k + \delta x_k) \approx \delta x_k^T \nabla^2 b_k^x (x_k)     \delta x_k + \\ 
    \delta x_k^T \nabla b_k^x (x_k) + b_k^x (x_k) 
\label{eqn:bf_quad}
\end{multline}
where $f_k^x$ is the constraint function at time step $k$. Ideally, a constraint function (e.g. log function) that imposes a hard constraint on obstacle avoidance is preferred over the exponential constraint due to its numerical nature (undefined for negative region). However, if trajectory initialization is infeasible, the optimization will fail because the region is undefined. In this case, we impose large weights on the soft constraints to ensure their dominance in the cost function.

One drawback of the CILQR is its need to have specific sets of weights tuned for specific sets of scenarios. For instance, using a set of weights tuned for an overtaking maneuver will always tend to encourage overtaking even if the target vehicle velocity is high. To improve the generality of the original CILQR and decrease the dependence on behavior-level planning, we propose analytical functions that take account of the state difference between the ego vehicle and the target vehicle along with the velocity of the target vehicle. These functions compute cost function weight adaptively and allow the planner to perform multiple tasks without switching weights.

\subsection{Adaptive Weight Tuning Function}
Intuitively, we want a lane keeping behavior when the forward target vehicle is far away from the ego vehicle irrespective of the speed of the target vehicle. When the distance gap closes, if the target vehicle is travelling close to the desired reference speed, a lane keeping behavior is desired. If the same target vehicle is travelling at a slow speed, a safe overtaking maneuver is desired. As a result, behaviors are closely related to target vehicle velocity and the difference between ego vehicle state and target vehicle state. 

In terms of cost function weights, the larger the velocity weight is, the more the planner encourages the ego vehicle to follow the reference velocity. The larger the reference weight is, the more the planner discourages the ego vehicle to explore options away from the reference trajectory.

Based on this observation, weights that affect the velocity and reference tracking term can be expressed in the following way:
\begin{align}
    \omega_{ref} &=  \frac{a_1}{v_{tgt}}\exp{\left(b_1*\Delta x\right)}  \label{eqn:w_fun_vel}\\
    \omega_{vel} &=  \frac{v_{tgt}}{a_2\exp{\left(b_2*\Delta x\right)}}  \label{eqn:w_fun_ref}
\end{align}
where $v_{tgt}$ is the target vehicle speed; $\Delta x$ is the distance difference between target and ego vehicles, and $a_1$, $a_2$, $b_1$, and $b_2$ are parameters to tune.

%Inclusions of these two terms in the cost function is shown in Equation \ref{eqn:w_fun_vel} and Equation \ref{eqn:w_fun_ref}.

\section{PROBLEM FORMULATION}\label{sec:problem_def}
\subsection{System Dynamics} 
\begin{figure}[h]
\includegraphics[width=0.45\textwidth]{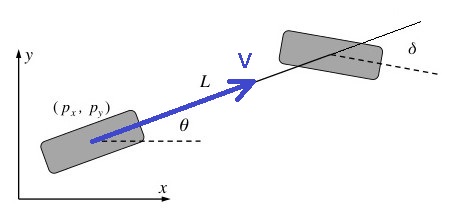}
\caption{Vehicle Kinematic Bicycle Model}
\label{fig:kinmodel}
\end{figure}

The kinematic bicycle model as shown in Figure \ref{fig:kinmodel} is used in this paper. The kinematic model considers a single pair of wheels at the center of front and rear axles instead of left and right pairs. The zero lateral slip assumption is made. The state consists of $ x = [p_x, p_y, v, \theta ]^T$ and the control input is $u = [ a, \delta]^T$,  where $p_x$, $p_y$ represent the position coordinates of the rear wheel, $v$ is the vehicle velocity, $\theta$ is the orientation of the vehicle, and $a$ and $\delta$ are the acceleration and steering angle of the vehicle respectively. Equation \ref{eqn:dyneqn} represents the transition function with non-zero steering angle. The horizontal increment is calculated by $\int_{0}^{l}\cos(\theta_t + \kappa s)ds$ and the vertical increment is calculated by $\int_{0}^{l}\sin(\theta_t + \kappa s)ds$. $\kappa = \frac{\tan(\delta)}{L}$ is the curvature and $l_t = v_t dt + \frac{1}{2}a* dt^2$ is the distance travelled at time t with discretization step $dt$. 
\begin{align}
    p_{x, t+1}  &= p_{x, t} + \frac{\sin(\theta_{t} + \kappa l) - \sin{\theta_t}}{\kappa} \nonumber \\
    p_{y, t+1}  &= p_{y, t} + \frac{\cos{\theta_t} - \cos(\theta_{t} + \kappa l)}{\kappa} \nonumber \\
    v_{t+1}     &= v_{t} + a*dt \nonumber \\
    \theta_{t+1}&= \theta_{t} + \kappa l \label{eqn:dyneqn}
\end{align}
However, when $\delta$ is zero, $\kappa$ will be zero, which will a cause numerical issue in Equation \ref{eqn:dyneqn}. Hence, a different update shown in Equation \ref{eqn:dyneqn_zero} is used. 
\begin{align}
    p_{x, t+1}  &= p_{x, t} + v_t dt \cos(\theta_t) \nonumber \\
    p_{y, t+1}  &= p_{y, t} + v_t dt \sin(\theta_t) \nonumber \\
    v_{t+1}     &= v_{t} + a*dt \nonumber \\
    \theta_{t+1}&= \theta_{t} \label{eqn:dyneqn_zero}
\end{align}

One thing to note is that in Figure \ref{fig:kinmodel} steering angle $\delta$ is present but the wheel slip angle is not. We assume there is no slipping for the vehicle, which eliminates the necessity for its inclusion in the model. 

\subsection{Objective Function}
The general definition of the objective function in Equation \ref{eqn:ILQR_obj_fun} can be made specific in Equation \ref{eqn:CILQR_obj_fun}. Each term in the summation represents the control effort cost, the reference trajectory tracking, the reference velocity tracking, and the constraints cost (i.e. control limits and obstacle avoidance). $c_N$ is the end state cost, which is the sum of the latter three terms as there is no control effort at the end state.
\begin{equation}
    J = \sum_{t = 0}^{N-1} \left( c_{t}^{u} + c_{t}^{ref} + c_{t}^{vel} + c_{t}^{con}\right) + c_N
    \label{eqn:CILQR_obj_fun}
\end{equation}

The reason we choose to separate the velocity term and the reference trajectory tracking term is the "pulling" effect stated in \cite{cILQR}. By taking only the closest point on the reference trajectory as the tracking point, we remove the time information otherwise encoded in pure-pursuit tracking. 

\subsubsection{control effort cost}
The acceleration and the steering angle are weighted differently as shown in \ref{eqn:c_u}, but they are both judged against zero, as we desire a minimum amount of control effort.  
\begin{equation}
    c_{t}^{u} = u_t^T\begin{bmatrix} w_{a} & \\ & w_{\delta}\end{bmatrix}u_t
    \label{eqn:c_u}
\end{equation}

\subsubsection{reference tracking and velocity cost}
The reference tracking term assigns a cost to each state based on its distance to the closest point on the reference trajectory. The closest distance can either be calculated against the nearest waypoint, or it can be calculated based on its projection on the reference trajectory. The velocity cost penalizes the ego vehicle for the difference between its velocity and the reference velocity. The combined cost is written in matrix form in Equation \ref{eqn:c_ref}. $\Delta x_t$ is the difference between the ego vehicle state and the reference state at time t. 
\begin{align}
    \Delta x_t &= x_t -  \begin{bmatrix} p_{x,t}^{ref} & p_{y,t}^{ref} &  v_{t}^{ref} & 0 \end{bmatrix}^T \nonumber \\
    c_{t}^{ref} + c_{t}^{vel} &= \Delta x^T\begin{bmatrix} w_{ref} &  &  & \\ & w_{ref} & & \\ & & w_{vel} & \\ & & & 0\end{bmatrix}\Delta x
    \label{eqn:c_ref}
\end{align}

\subsubsection{constraint cost}
All inequality constraints can be expressed in a negative null form shown in Equation \ref{eqn:negative_null} in which $x_{lim}$ is the maximum or minimum boundary value and $f(x)$ is some sort of function on the decision variable.

\begin{equation}
    g(x) = x_{lim} - f(x) \leq 0
    \label{eqn:negative_null}
\end{equation}

For linear constraints like acceleration and steering limits, we can write them as for instance: $$g(u) = u - u_{max} \leq 0$$ Then, a barrier function can be used as stated in Equation \ref{eqn:bf}. 

The ego vehicle is represented by two circles centered at the middle of the rear and front axle. For the obstacle avoidance term, we can either use the closest distance between two vehicles and set a minimum safety distance or use a geometric collision check as the inequality constraint. Since we need to investigate the worst performance of different planning strategies later, minimum safety distance is set to zero and the problem is formulated as a geometry-based cost function. Obstacles are formulated as ellipses with major and minor axes adjusted for ego vehicle geometry for computation ease when taking the Jacobian and the Hessian of the cost function for the LQR backward pass. The over-fitting is the minimum area ellipse and the inequality constraint is shown in Equation \ref{eqn:ellipse}. $\theta_t^{obs}$ is the heading angle of the obstacle at time $t$, $a$ and $b$ are semi major and minor axes' lengths. $\Delta x_t$ is the position difference between the ego vehicle and the obstacle.
\begin{align}
    R_t &= \begin{bmatrix} \cos(\theta_t^{obs}) & -\sin(\theta_t^{obs}) \\
                           \sin(\theta_t^{obs}) & \cos(\theta_t^{obs})
            \end{bmatrix} \nonumber \\
    A_t &= R_t\begin{bmatrix} \frac{1}{a^2} &  \\
                                             & \frac{1}{b^2}
            \end{bmatrix}R_t^T \\
    g(x_t) &= 1 - \Delta x_t^T * A_t * \Delta x_t \leq 0\label{eqn:ellipse}
\end{align}
For short-term prediction, since the output of the reachability analysis is a sequence of rectangular bounding boxes, each box essentially represents the upper bound and lower bound of the mass point's future $(x, y)$ position. We further inflate the boxes into ellipses considering the vehicle's shape. These ellipses will be addressed as the obstacles over the short-term prediction horizon.

The same inequality constraint needs to be repeated for the front center of the ego vehicle as well. Reformulated inequality constraints can then be used after linearization and barrier function transformation.

% \begin{figure}[h]
% \centering
% \includegraphics[width=5cm]{figures/car_two_circ.png}
% \caption{Two-Circle Representation of the Ego Vehicle}
% \label{fig:two_circ}
% \end{figure}

\section{EXPERIMENTAL RESULTS} \label{sec:result}
In this section, the environment is set up to have two lanes. The top lane ranges from $y = 0\ m$ to $y = 6\ m$ with the center of the lane located at $y = 3\ m$. The bottom lane ranges from $y = 0\ m$ to $y = -6\ m$ with the center of the lane located at $y = -3\ m$. Solid black lines and red dashed lines represent lane separators and lane centers, respectively. The reference velocity for all test cases is set to 15 m/s (roughly 35 mph), which is a common speed limit. The acceleration is limited to $[-4, 6] m/s^2$ and the steering angle is limited to $\pm30^\circ$.

The ego vehicle is pictured in blue with its transparency decreased over time (i.e. white is the frame at $t_0$, the less transparent the color is the later a frame is in time). Target vehicles are depicted in red and the transparency works the same way. The finite planning horizon is set to 4 seconds with a discretization of 0.1 seconds. 

Three subsections are presented below:
\begin{enumerate}
    \item The ego vehicle overtakes a slow-travelling target vehicle in front with an opposing vehicle travelling in the opposite lane. Also, comparisons against CILQR with set weights are conducted in this section.
    \item The ego vehicle performs an evasive maneuver against an incoming cut-in vehicle and then performs a car-following behavior behind the target vehicle.
    \item The target vehicle speeds up unexpectedly during overtaking maneuver, which shows the importance of the short-term prediction comparing to ones using deterministic trajectories \cite{cILQR, cILQR2}.
\end{enumerate}

% In the first subsection, the ego vehicle overtakes a slow-travelling target vehicle in front. This case shows the basic functionality of the framework we propose. We also performed some comparisons between the proposed framework and the original CILQR relating to the weight tuning effects. We define a transient behavior, in which the ego vehicle performs a hesitating maneuver with a sinusoidal like trajectory. The transient behavior is potentially dangerous because these sinusoiodal trajectories have significant deviation from the lane center and could pose collision hazards to surrounding vehicles. The percentage of the transient behavior appearance is studied as a benchmark of the effectiveness of the adaptive weight tuning. 

% In the second subsection, the ego vehicle performs an evasive maneuver against an incoming cut-in vehicle and then performs a car-following behavior behind the target vehicle. The adaptive weight tuning is also working in these two scenarios. Using the same set of weights, the ego vehicle performed both a overtaking and car-following maneuver compared to the requirement of two distinct sets of weights for the one experimented with \cite{cILQR}. 

% The third subsection investigates the importance of the short-term reachability by comparing against a planner using only a long-term prediction. Note that all test cases use short-term and long-term prediction instead of the deterministic trajectories or persistent prediction for the target vehicle in the literature \cite{cILQR,cILQR2,chen2019autonomous}.

\subsection{General Performance Evaluation}
\begin{figure}[H]
\centering
\includegraphics[width=0.5\textwidth]{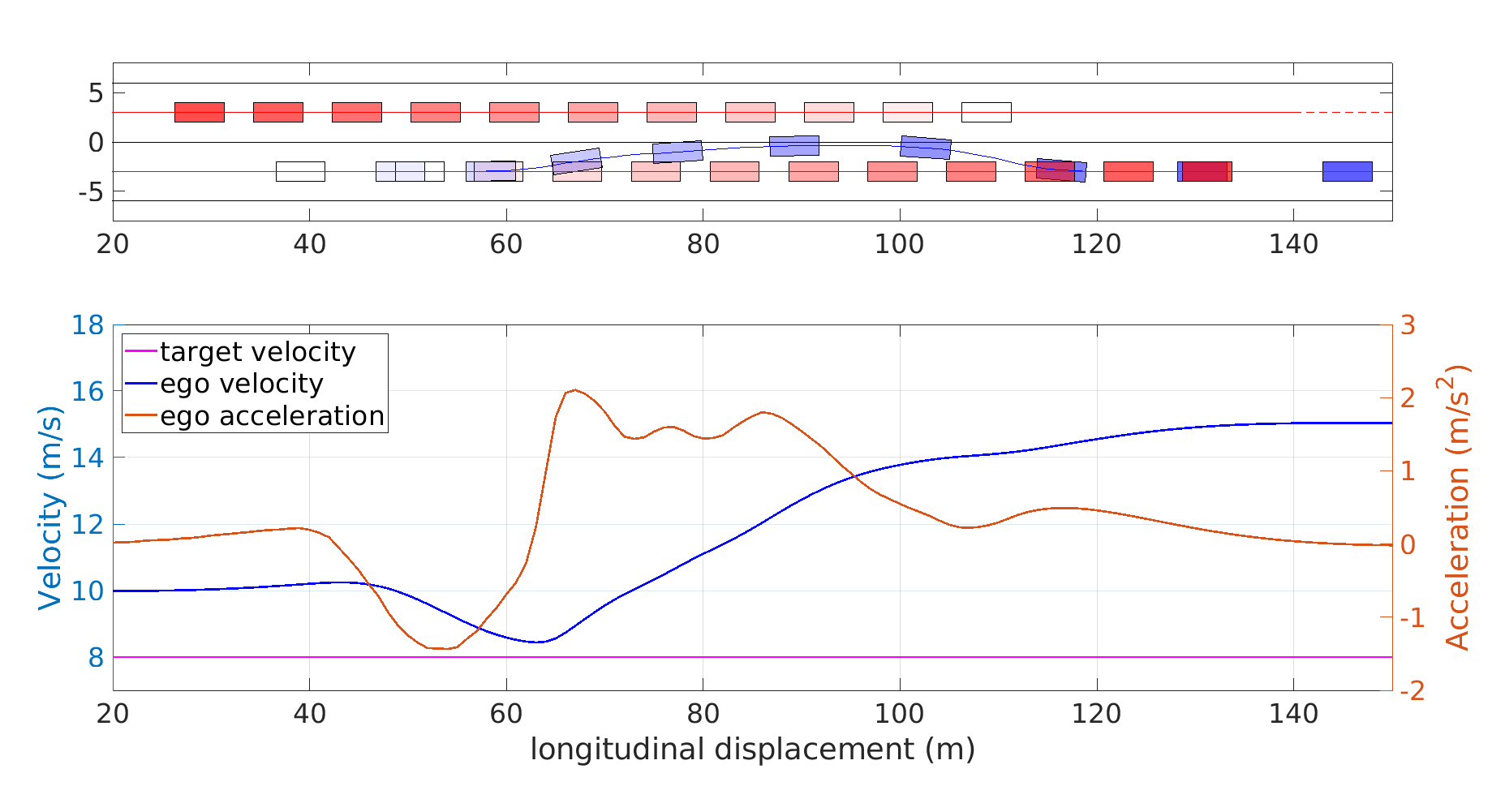}
\caption{Critical Section of Overtaking a Slow Front Vehicle Travelling 8m/s}
\label{fig:A1_fig}
\end{figure}

%Overtaking a slow target vehicle is important for arriving at the destination efficiently.
In this scenario, a slow vehicle is initially 20 meters ahead of the ego vehicle in the adjacent lane and it is travelling at 8 m/s.  There is an incoming vehicle travelling in the opposing lane and the task is to safely overtake the front vehicle. As seen in the velocity profile of Figure \ref{fig:A1_fig}, the ego vehicle first slows down behind the target vehicle to ensure safety and then pulls over for the overtake while avoiding the incoming vehicle. During the overtaking maneuver, the ego vehicle smoothly accelerates to the reference velocity of 15 m/s. This test case is used to show the basic ability of the planning framework we propose. This is an extreme case designed to test the performance of the planner while the ideal action is probably overtaking after the opposing vehicle has passed.

For comparisons with the original CILQR method, we use the transient behavior percentage as a performance benchmark. We define the transient behavior as behaviors in which the ego vehicle performs a hesitating maneuver with a sinusoidal like trajectory. The transient behavior is potentially dangerous because these sinusoidal trajectories have significant deviations from the lane center and could pose collision hazards to surrounding vehicles especially when the surrounding vehicles need to perform a lane change into the ego vehicle's lane. A set of weights is tuned for overtaking in the CILQR method and a set of adaptive weighting function parameters is tuned to encourage overtaking against target vehicles with velocity less than 13 m/s. The results are listed in Table \ref{tab:Comparison}. For CILQR, when the target vehicle velocity exceeds 12 m/s, all planned trajectories are transient ones. For the method we propose, there are no transient behaviors observed and planned trajectories are either car following ones or overtaking ones. To achieve the two different behaviors with CILQR, two sets of weights need to be tuned manually, which is avoided in the proposed method. This indicates that the framework we propose is a safer planner with increased generality.

\begin{table}[H]
\centering
\caption{CILQR vs. Adaptive Weight Tuning Performance Comparison for Target Vehicle Speed Range from 8 m/s to 14.9 m/s}
\scriptsize 
\begin{tabular}{|l|l|l|l|l|l|l|}
\hline
Method  & \multicolumn{3}{c|}{CILQR}  & \multicolumn{3}{c|}{Adaptive Weight Tuning}  \\ \hline
Behaviors    & \begin{tabular}[c]{@{}l@{}}Over-\\ take\end{tabular} & \begin{tabular}[c]{@{}l@{}}Tran-\\ sient\end{tabular} & \begin{tabular}[c]{@{}l@{}}Lane-\\ keep\end{tabular} & \begin{tabular}[c]{@{}l@{}}Over\\ -take\end{tabular} & \begin{tabular}[c]{@{}l@{}}Tran-\\ sient\end{tabular} & \begin{tabular}[c]{@{}l@{}}Lane-\\ keep\end{tabular} \\ \hline
occurrences                                                  & 48                                                   & 22                                                    & 0                                                    & 56                                                   & 0                                                     & 14                                                   \\ \hline
\begin{tabular}[c]{@{}l@{}}percentage \\ (\%)\end{tabular}   & 68.57                                                & 31.43                                                 & 0                                                    & 80.00                                                & 0                                                     & 20.00                                                \\ \hline
\begin{tabular}[c]{@{}l@{}}speed range \\ (m/s)\end{tabular} & 8-11.9                                               & 12-14.9                                               & 0                                                    & 8-13.5                                               & 0                                                     & 13.6-14.9                                            \\ \hline
\end{tabular}
\label{tab:Comparison}
\end{table}

\subsection{Adjacent Vehicle Violently Cuts in}
Due to blind spot awareness, drivers from adjacent lanes might perform a lane change maneuver without environmental awareness. This is a dangerous situation, as the target vehicle is unlikely to yield and distance between two vehicles is very close, which limits the reaction time.

\begin{figure}[h]
\includegraphics[width=0.5\textwidth]{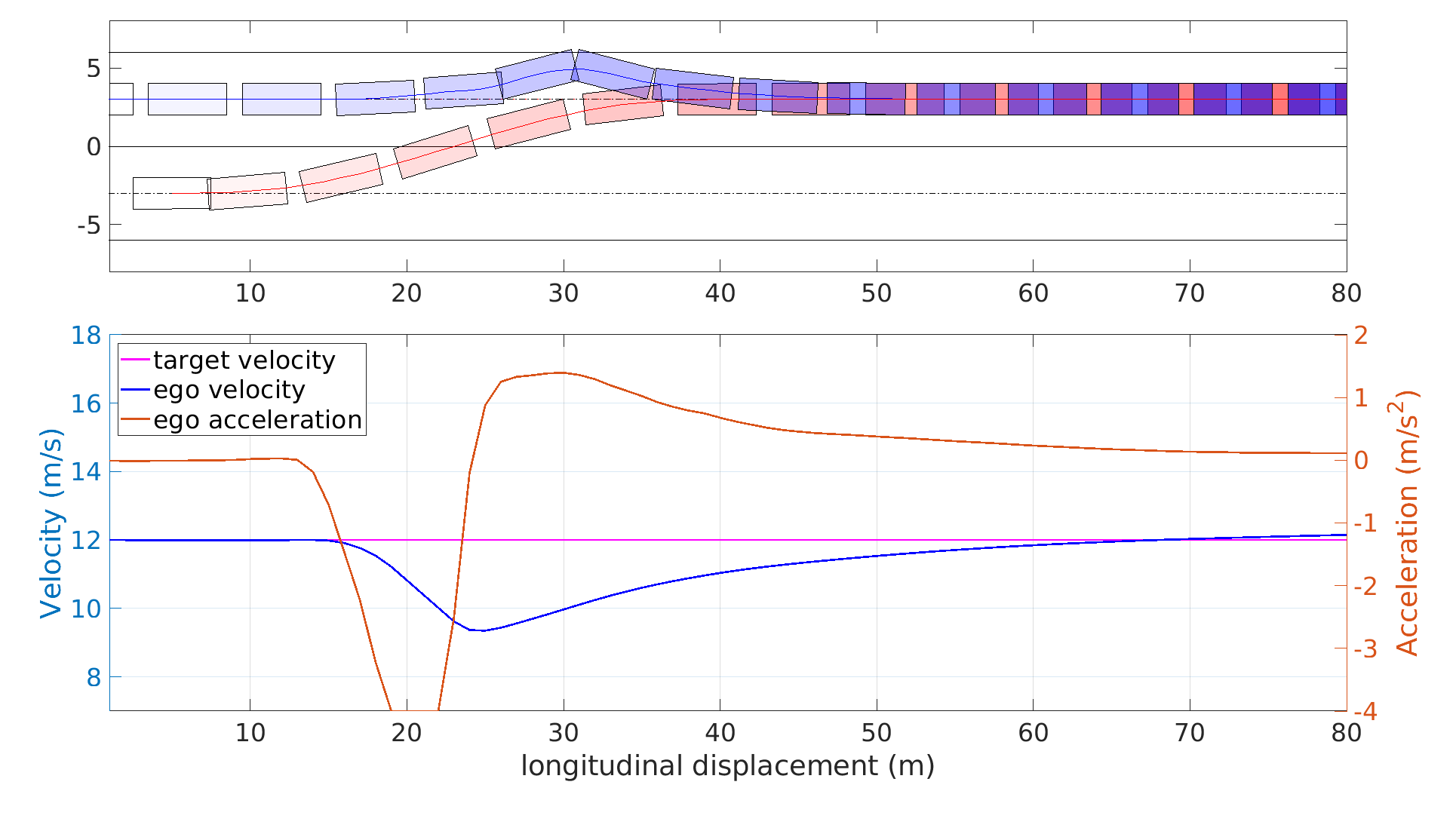}
\caption{Target Vehicle Travelling 12 m/s Violently Cuts into the Ego Lane}
\label{fig:D1_fig}
\end{figure}

In this scenario, a target vehicle that is initially 5 meters ahead of the ego vehicle in the adjacent lane violently cuts into the ego vehicle's lane, as shown in Figure \ref{fig:D1_fig}. Trajectories of the ego vehicle and the target vehicle are depicted in solid blue and red lines in the top figure correspondingly. The bottom plot shows the velocity and the acceleration profile. The blue and the purple lines are velocities of the ego vehicle and the target vehicle, respectively. The red line is the acceleration of the ego vehicle. The target trajectory is generated by a fifth-order polynomial fitting. It completes the lane changing maneuver in 3 seconds. 

Around $x = 30m$, the planner reacts to the cut-in vehicle. It slows down and performs a yielding turn to avoid the incoming target vehicle. As the target vehicle passes by, it then follows the target vehicle using a car-following behavior. We further test for the safety margin of the planner with decreased cut-in vehicle velocity and shorter reaction time. The failure cases require the adjacent vehicle to travel at a speed around 8 m/s and finish the cut-in maneuver within 0.3 to 0.4 seconds. As this minimum safety criterion is hardly kinematically feasible, the safety of the planner is guaranteed.

\begin{figure}[h]
\centering
\includegraphics[width=0.45\textwidth]{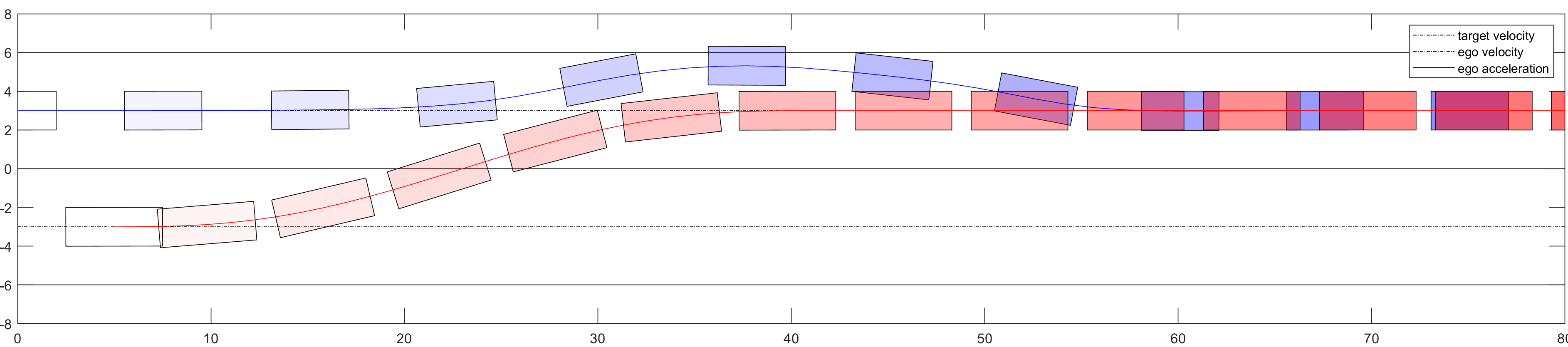}
\caption{Cut-in Scenario Time Optimal Solution}
\label{fig:D1_fig_optimal}
\end{figure}

We then compare the planned trajectory against a planner using deterministic knowledge of the obstacle. In this planner, we used the obstacle future trajectory directly as the prediction information to generate an optimal solution as the baseline. The baseline optimal solution, shown in Figure \ref{fig:D1_fig_optimal}, is to speed up and go around the incoming target vehicle. Compared to the optimal solution, we can see that the planner we proposed is more conservative in terms of whether to overtake the target vehicle when it cuts in. This is safer than surpassing the target vehicle in reality.

\subsection{Target Vehicle Suddenly Accelerates during Overtaking Maneuver}
\begin{figure}
\centering
\subfloat[Unsafe Trajectory and Velocity Profile.\label{fig:B2_no_reach_general}]{\includegraphics[width=8cm]{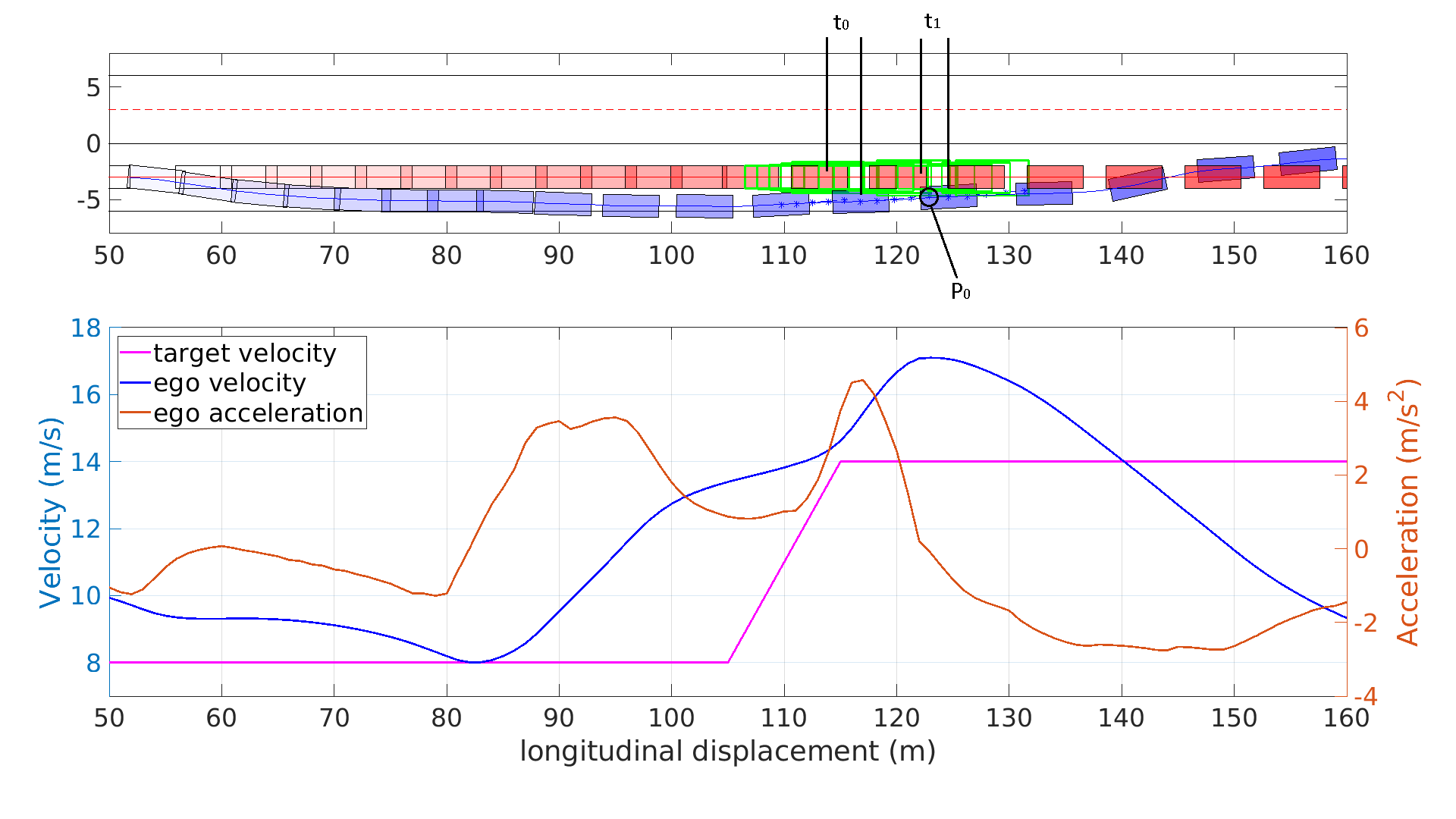}}

\subfloat[Enlarged Critical Section. Same time frames of two cars are aligned with pairwise vertical black lines.\label{fig:B2_no_reach_detailed}]{\includegraphics[width=8cm]{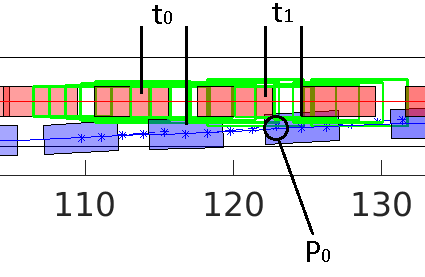}}
\caption{Unsafe Corner Case Using Only Long-Term Prediction} 
\label{fig:B2_no_reach} 
\end{figure}

In this scenario, the target vehicle that is being overtaken suddenly starts to accelerate to avoid being overtaken. It imitates an aggressive style of driving. Green boxes are reachability analysis at a certain time step. It predicts 0.5 seconds ahead of time with a step of 0.1 seconds each. They are also drawn in Figure \ref{fig:B2_no_reach} for visualization purposes against planned trajectory for 0.5 seconds (shown in blue asteroid lines). 

% For the case in which reachability analysis is used, the ego vehicle reacts to the sudden acceleration of the target vehicle and yields to the target vehicle as shown in Figure \ref{fig:B2_reach}. For the case shown in Figure \ref{fig:B2_no_reach} in which only the long-term prediction is used, the ego vehicle does not recognize the sudden acceleration in time. In the second frame with reachability set drawn in Figure \ref{fig:B2_no_reach} (around $x = 115m$), the planned trajectory for the frame intersects with the target reachability set. It indicates the planned trajectory may enter the reachable region of the target vehicle in the short-term horizon. We observe that it results in a collision in the succeeding step. One thing to note is that to examine the worst performance of two motion planning frameworks, the obstacle avoidance constraint is relaxed from keeping a minimum distance $r_{safe}$ to a zero distance-keeping so that the performance is not affected by another parameter.
%\label{fig:B2_no_reach_general}
In Figure \ref{fig:B2_no_reach}, only the long-term prediction is used for the whole 4s planning. Note again that the short-term prediction is disabled in this case and the visible green reachable set is only used for investigating its capability of safety guarantee. The target vehicle pictured in red starts to accelerate at around 105 meters mark in Figure \ref{fig:B2_no_reach_general}. At time $t_0$, the planned trajectory that is 0.5 seconds ahead intersects with the reachable set, which is labeled as $p_0$ in Figure \ref{fig:B2_no_reach_detailed}. It demonstrates that the planned trajectory is not guaranteed to be collision-free. Not surprisingly, at the succeeding time step $t_1$, the ego vehicle collides with the target vehicle because the long term planner does not react timely to the acceleration.

% \begin{figure}[h]
% \includegraphics[width=0.5\textwidth]{IROS_Figure/B2_paper_no_reach_ver5.png}
% \caption{Unsafe Corner Case Using Only Long-Term Prediction}
% \label{fig:B2_no_reach}
% \end{figure}

%\begin{figure}[h]
%    \begin{subfigure}{0.5\textwidth}
%        \centering
%        % include first image
%        \includegraphics[width=\linewidth]{IROS_Figure/B2_paper_no_reach_ver5.png}
%        \caption{Unsafe Trajectory and Velocity Profile}
%        \label{fig:B2_no_reach_general}
%    \end{subfigure}
%
%%\newline
%
%    \begin{subfigure}{0.5\textwidth}
%        \centering
%        % include first image
%        \includegraphics[width=0.8\linewidth]{IROS_Figure/B2_paper_no_reach_enlarged.png}  
%        \caption{Enlarged Critical Section. Same time frames of two cars are aligned with pairwise vertical black lines.}
%        \label{fig:B2_no_reach_detailed}
%    \end{subfigure}
%    
%
%    
%    
%\caption{Unsafe Corner Case Using Only Long-Term Prediction}
%\label{fig:B2_no_reach}
%% \vspace{-0.8cm}
%\end{figure}

\begin{figure}[h]
\includegraphics[width=0.5\textwidth]{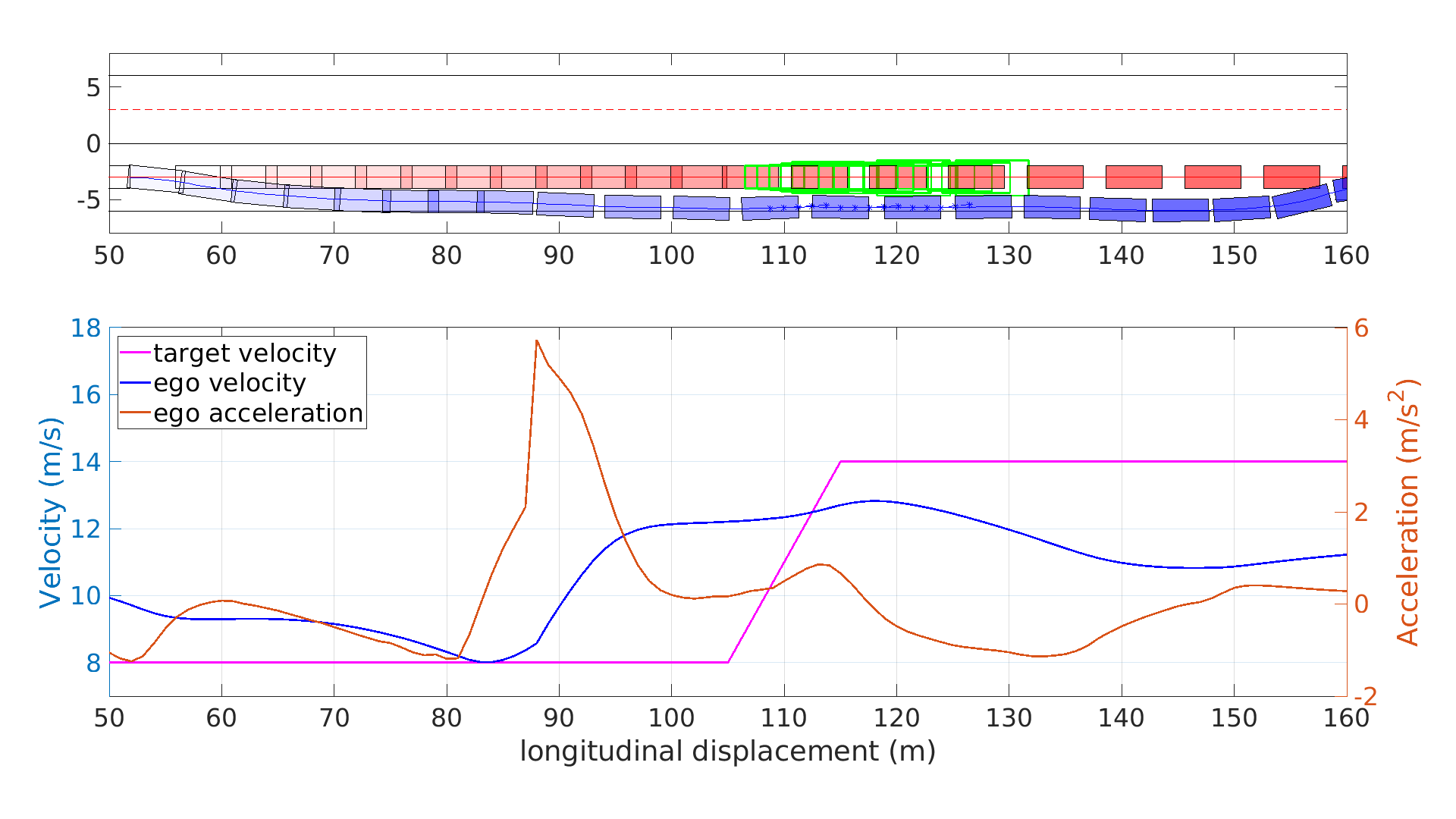}
\caption{Safe Planning Using Reachability for the Short-Term Prediction}
\label{fig:B2_reach}
\end{figure}

In Figure \ref{fig:B2_reach}, all planned trajectories at each time step take reachability analysis into account and the planner avoids the collision case in Figure \ref{fig:B2_no_reach_general}.  One thing to note is that to examine the worst performance of two motion planning frameworks, the obstacle avoidance constraint is relaxed from keeping a minimum distance $r_{safe}$ to a zero distance-keeping so that the performance is not affected by another parameter. As a result, for extreme cases, the short-term prediction guarantees the planned trajectory robust and safe enough against sudden target vehicle movement changes. The reason is that planning to avoid the reachable region allows the target vehicle to perform any non-deterministic but kinematically feasible behaviors over the short-term horizon.

The average loop time for these experiments is 60 ms with a standard deviation of 25 ms. The simulation is written in Matlab and runs on a laptop with a 2.80GHz Intel Core i7-7700HQ CPU. With possible implementation in C++, the performance can be further accelerated.

\section{CONCLUSION}\label{sec:conclusion}
In this paper, the existing CILQR framework is improved in two respects. We first added uncertainty awareness to the framework and made it more robust to sudden target vehicle maneuvers. The safety can be guaranteed because reachability is essentially an enclosure of all possible trajectories of the target vehicle over the short-term horizon. The long-term prediction is utilized to avoid over-conservativness. Secondly, an adaptive weight calculation function is added to the algorithm. It allows the planner to perform multiple maneuvers without tuning for different cost function weights.

There still exist several interesting problems. The first is the tradeoff between the hard obstacle avoidance constraint and the numerical stability when incorporating the barrier function. The second problem is the inclusion of jerk within the cost function for the purpose of comfort. One possible method is to expand the current state-space model for the inclusion of the jerk or introduce jerk constraints on the acceleration.

\section{ACKNOWLEDGMENT}
This material is based upon work supported by the United States Air Force and DARPA under Contract No. FA8750-18-C-0092. Any opinions, findings and conclusions or recommendations expressed in this material are those of the author(s) and do not necessarily reflect the views of the United States Air Force and DARPA.
%%%%%%%%%%%%%%%%%%%%%%%%%%%%%%%%%%%%%%%%%%%%%%%%%%%%%%%%%%%%%%%%%%%%%%%%%%%%%%%%

%%%%%%%%%%%%%%%%%%%%%%%%%%%%%%%%%%%%%%%%%%%%%%%%%%%%%%%%%%%%%%%%%%%%%%%%%%%%%%%%

%%%%%%%%%%%%%%%%%%%%%%%%%%%%%%%%%%%%%%%%%%%%%%%%%%%%%%%%%%%%%%%%%%%%%%%%%%%%%%%%
% \section*{APPENDIX}

\bibliographystyle{IEEEtran}
% \bibliography{IEEEabrv,IEEEexample}
\bibliography{citation}

\end{document}